\documentclass{article}

\usepackage[noend]{algorithmic}

\usepackage[accepted]{icml2022}

\usepackage{microtype}
\usepackage{graphicx}
\usepackage{subfigure}
\usepackage{booktabs} 
\usepackage{graphicx}
\usepackage{amsmath}
\usepackage{xcolor}
\usepackage{amsmath}
\usepackage{tabularx} 
\usepackage{amsfonts}
\usepackage{makecell}

\newenvironment{itemize*}%
  {\vspace{-\topsep}
   \begin{itemize}%
     \setlength{\itemsep}{0pt}}%
  {\end{itemize}
   \vspace{-\topsep}}

\newenvironment{enumerate*}%
  {\vspace{-\topsep}
   \begin{enumerate}%
    \setlength{\itemsep}{0pt}}%
  {\end{enumerate}
   \vspace{-\topsep}}

\usepackage{hyperref}

\renewcommand{\algorithmiccomment}[1]{{#1}}

\icmltitlerunning{Rich Feature Construction for the Optimization-Generalization Dilemma}

\begin{document}

\twocolumn[
\icmltitle{Rich Feature Construction for the Optimization-Generalization Dilemma}


\icmlsetsymbol{equal}{*}

\begin{icmlauthorlist}
\icmlauthor{Jianyu Zhang}{nyu}
\icmlauthor{David Lopez-Paz}{fbpar}
\icmlauthor{L\'eon Bottou}{fbnyc,nyu}
\end{icmlauthorlist}

\icmlaffiliation{nyu}{New York University, New York, NY, USA.}
\icmlaffiliation{fbpar}{Facebook AI Research, Paris, France.}
\icmlaffiliation{fbnyc}{Facebook AI Research, New York, NY, USA.}

\icmlcorrespondingauthor{Jianyu Zhang}{jianyu@nyu.edu}

\icmlkeywords{Out-of-distribution generalization, invariance}

\vskip 0.3in
]



\printAffiliationsAndNotice{}  

\begin{abstract}


There often is a dilemma between ease of optimization and robust out-of-distribution (OoD) generalization. For instance, many OoD methods rely on penalty terms whose optimization is challenging. They are either too strong to optimize reliably or too weak to achieve their goals. 

We propose to initialize the networks with a rich representation containing a palette of potentially useful features, ready to be used by even simple models. On the one hand, a rich representation provides a good initialization for the optimizer. On the other hand, it also provides an inductive bias that helps OoD generalization. Such a representation is constructed with the Rich Feature Construction (RFC) algorithm, also called the \emph{Bonsai} algorithm,{\footnotemark} which consists of a succession of training episodes. During \emph{discovery episodes}, we craft a multi-objective optimization criterion and its associated datasets in a manner that prevents the network from using the features constructed in the previous iterations. During \emph{synthesis episodes}, we use knowledge distillation to force the network to simultaneously represent all the previously discovered features. 

Initializing the networks with Bonsai representations consistently helps six OoD methods achieve top performance on \textsc{ColoredMNIST} benchmark \cite{irm}. The same technique substantially outperforms comparable results on the Wilds \textsc{Camelyon17} task \cite{wilds2021}, eliminates the high result variance that plagues other methods, and makes hyperparameter tuning and model selection more reliable.
\end{abstract}
\vfill
~
\footnotetext{Bonsai, also known as \emph{penjing} (tray planting), refers to the art of growing small trees in trays using obsessive trimming techniques to impede their growth and produce miniature versions of real-life trees. Likewise, the RFC algorithm impedes the learning process in order to obtain diverse representations.}

\section{Introduction}
\label{sec:introduction}

The interplay of optimization and generalization plays a crucial role in deep learning. It also changes nature when we focus on out-of-distribution (OoD) generalization, that is when training and testing sets are no longer assumed to follow the same distribution.

Simple optimization algorithms are surprisingly able to find uneventful descent paths in the non-convex cost landscape of deep learning networks \cite{gu2021domainfree, Sagun2017}. However they also tend to construct features that capture spurious correlations~\citep{szegedy-2014, beery2018recognition, irm, ilyas-madry-2019}. Several recent papers propose to work around this problem by leveraging multiple training sets that illustrate the possible changes in distribution. The training algorithm must then satisfy certain constraints across training sets, usually enforced with additional penalty terms  \citep{irm,iga,vrex,sd,gm,rame2021fishr, wald2021calibration}. The resulting optimization problem often turns substantially more challenging than the simple empirical risk minimization (ERM). 
In practice, it is often necessary to schedule the penalty hyper-parameters in a manner that weakens them so much that they no longer enforce the intended constraints. As a result, when one initializes such a method with the correct solution, the training process deviates from the constraint set and finds inferior solutions (Figure \ref{fig:generalization_difficulty}).

We propose in this paper to work around the difficulties of the optimization problem by first obtaining a representation of the input patterns that contains a broad diversity of potentially useful features. Both the ERM and OoD methods can then easily pick the most useful features, according to the chosen training cost and constraints, in a much easier way. The Rich Feature Construction (RFC) algorithm, informally called the Bonsai algorithm, (Section~\ref{sec:rfc}) consists of a succession of training episodes. During the \emph{discovery episodes}, we craft a multi-objective optimization criterion that prevents the network from using the features constructed in the previous steps.
During the \emph{synthesis episodes}, we force the final representation to simultaneously represent all the previously identified features.

On the common out-of-distribution (OoD) \textsc{ColoredMNIST} task~\citep{irm}, we show that initializing networks with Bonsai representations consistently helps six state-of-the-art OoD learning algorithms learn the robust feature and disregard the spurious one. The same method also performs well on a modified task, \textsc{ReverseColoredMNIST}, in which the robust feature is made more predictive than the spurious feature. Such a modification breaks all methods that aim for the 2nd easiest-to-find features~\cite{lff, liu2021just, Bao2021}. Finally, we evaluate Bonsai initialization on the real-world \textsc{camelyon17} task \cite{wilds2021} and show that it not only helps OoD and ERM methods match or exceed the best published results, but also facilitates the hyper-parameter tuning and model selection.\footnote{Code for replicating these experiments is publicly available at \url{https://github.com/TjuJianyu/RFC/}.\hfill}

\section{Related Work}

\subsection{Leveraging multiple training environments}
In order to achieve a good performance on testing data that follows a different distribution from the training data, many OoD methods assume access to multiple training sets, or environments, whose different distributions illustrate a range of potential distribution changes. 
One possible direction consists of learning a representation such that the optimal classifier built on top is the same for all training environments: IRMv1/IRM \cite{irm}, MAML-IRM \cite{mamlirm}, CLOvE \cite{wald2021calibration}. Another line of work introduces gradient alignment constraints across training environments using dot-product (Fish \cite{gm}), squared distance of gradients (IGA \cite{iga}), or squared distance of gradients variance (Fishr \cite{rame2021fishr}).
Methods such as vREx \cite{vrex} and GroupDRO \cite{groupDRO} aim at finding a solution that performs equally well across training environments. 

\subsection{Facilitating the optimization}

In contrast, the SD method \cite{sd} relies on a single training set but fights the "gradient starvation" phenomenon that prevents the training algorithm from finding robust features even though they are assumed more predictive on the training set than easier-to-find spurious features.

\subsection{Aiming for the second easiest representation}
\label{sec:relatedwork_discover_2nd_feat}

Closely related to the RFC \emph{discovery episodes}, several methods seek the \emph{second easiest-to-find representation}, either by reweighing the dataset \citep{liu2021just, lff} or with distribution robust optimization, \citep{Bao2021, ahmed2020systematic, creager2021environment}. The main drawback of these methods is the assumption that the second-easiest representation is the correct one. This happens to be true in benchmarks such as \textsc{ColoredMNIST} which are designed to frustrate ERM. However, these methods fail on simpler tasks that do not follow the assumption (Section \ref{sec:2nd_easy_feature}). 
    
\subsection{Making use of diverse features}
Closely related to the RFC \emph{synthesis episodes}, other methods attempt to steer the training process towards constructing a diversity of features: RSC \citep{rsc} masks out features associated with large gradients to force the last layer to pick additional features; DiverseModel \citep{Teney2021} constructs multiple classifiers on top of a given representation (e.g. ImageNet pre-trained) with a penalty that minimizes the alignment of gradients across classifiers. The drawback of these approaches is that they work best when starting from an existing portfolio of diverse features.

\section{The optimization-generalization dilemma}
\label{sec:spurious_tour}

This section presents experiments that illustrate the
optimization-generalization dilemma that plagues OoD methods.
All these experiments are carried out on the \textsc{ColorMNIST}
task \citep{irm}. In this task, the relation between the robust feature (the digit class) and output label is invariant in all training and testing  environments. In contrast, although the spurious feature (the digit color) is more predictive on the training environments, its relation with the output labels is not invariant across environments.
We report results on a variety of published OoD algorithms: IRMv1 \cite{irm}, Fish \cite{gm}, IGA \cite{iga}, vREx \cite{vrex}, Spectral Decoupling (SD) \cite{sd}, Fishr \cite{rame2021fishr}, RSC \cite{rsc}, LfF \cite{lff}, and CLOvE \cite{wald2021calibration}. We do not report results on MAML-IRM \cite{mamlirm} because it is equivalent to Fish+vREx, and we do not report results on GroupDRO \citep{groupDRO} because it performs like vREx (see Appendix \ref{apdix:mamlirm=vrex+gm} and \ref{apdix:groupdro_vrex_inter_extra} for details).

\subsection{OoD penalties make the optimization challenging}
\label{sec:importance_of_initialization_ood}

\begin{figure}
    \centering
    \includegraphics[width=0.48\textwidth]{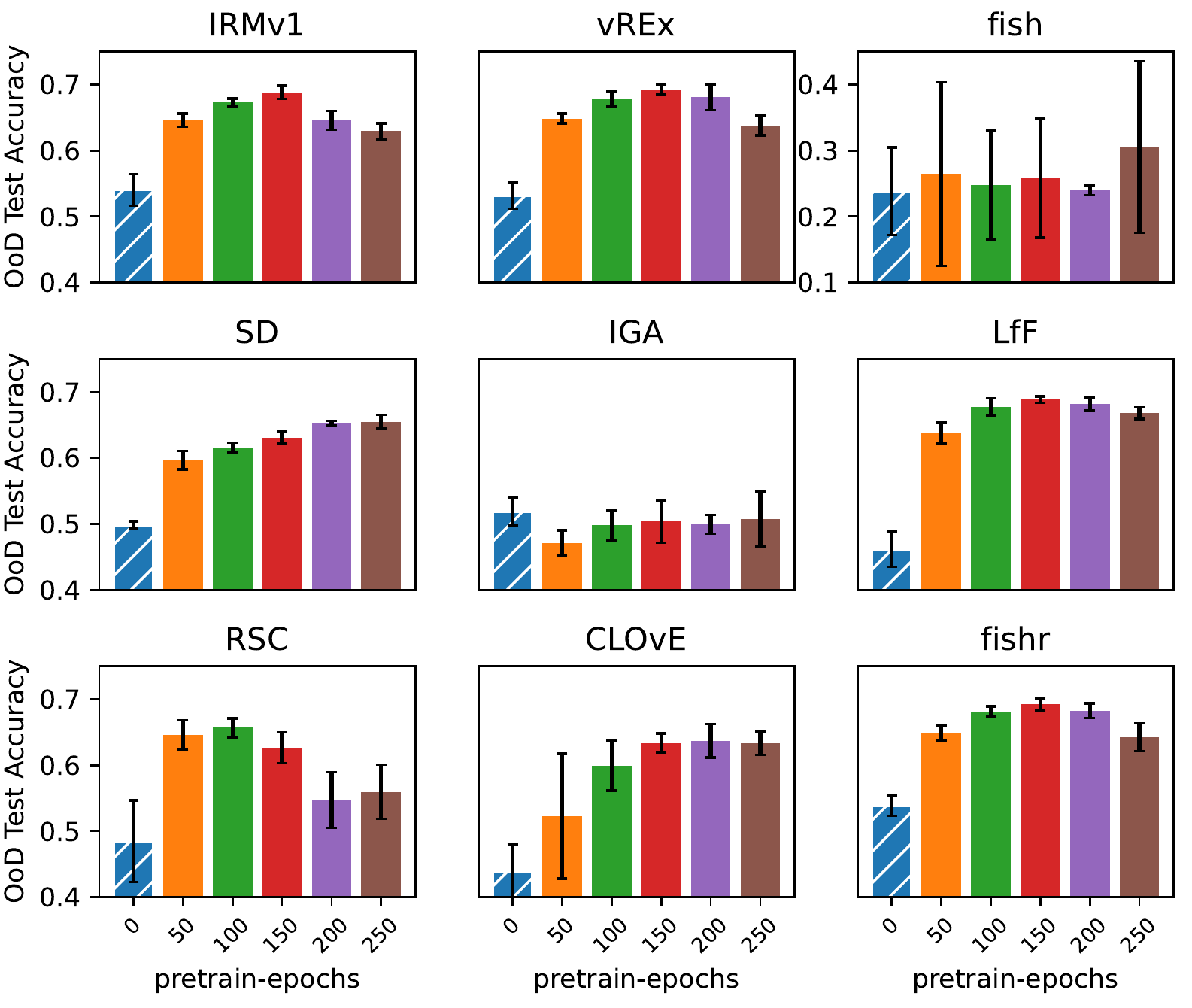}
    \caption{Test performance of nine penalized OoD methods as a function of the number of epochs used to pre-train the neural network with ERM. The final OoD testing performance is very dependent on choosing the right number of pretraining epochs, illustrating the challenges of these optimization problems.}
    \label{fig:importance_of_rep}
\end{figure}

Because their optimization is difficult, most authors recommend to pre-train the network with ERM before applying their OoD method. 
Figure~\ref{fig:importance_of_rep} shows the final OoD test performance of models trained with each method as a function of the number $n_p\in\{0, 50, 100, 150, 200, 250\}$ of ERM pretraining epochs. During the execution of the OoD algorithm, we choose one of five penalty weights and select the best early-stopping epoch by directly peeking at the OoD test performance. All other hyper-parameters are copied from the \textsc{ColorMNIST} task \cite{irm}.  Appendix~\ref{apdix:train_details_colorminst} discusses these experiments with further details.

Figure~\ref{fig:importance_of_rep} shows that optimizing from
a random initialization (blue bars, 0 pretraining epochs) fails for all nine algorithms and all five penalty weights. Although pretraining with ERM helps, the final performance of the competitive algorithms depends on the number of pretraining epochs in rather inconsistent. Too much pretraining can cause performance drops in excess of 20\%. Even when one guesses the right amount of pretraining, the final performance comes short of the oracle performance ($0.721\pm0.002$) achieved by a network that is trained only on the robust feature.

We also showcase the optimization difficulty of several OoD methods from a loss landscape's view on a low-dimensional case. See Appendix \ref{apdix:lowdim_losslandscape} for details. 

\subsection{OoD penalties do not enforce the constraints}
\label{sec:generalization_difficulty}

The previous section shows that the penalties introduced by these OoD methods are too strong to allow reliable optimization. We now show that they are also too weak to enforce the constraints they are meant to enforce.

To substantiate this assertion, we initialize a network with the correct solution, that is, the solution obtained by training the network on a variant of the \textsc{ColorMNIST} dataset in which the
spurious color feature was removed. In order to keep the network from deviating from the target constraint, we use the largest penalty weight in the search space in each OoD method. We do not report results on the Fish method because it failed to learn the task. We do not report on RSC and LfF because their test accuracy drops too fast.

\begin{figure}
    \centering
    \includegraphics[width=0.473\textwidth]{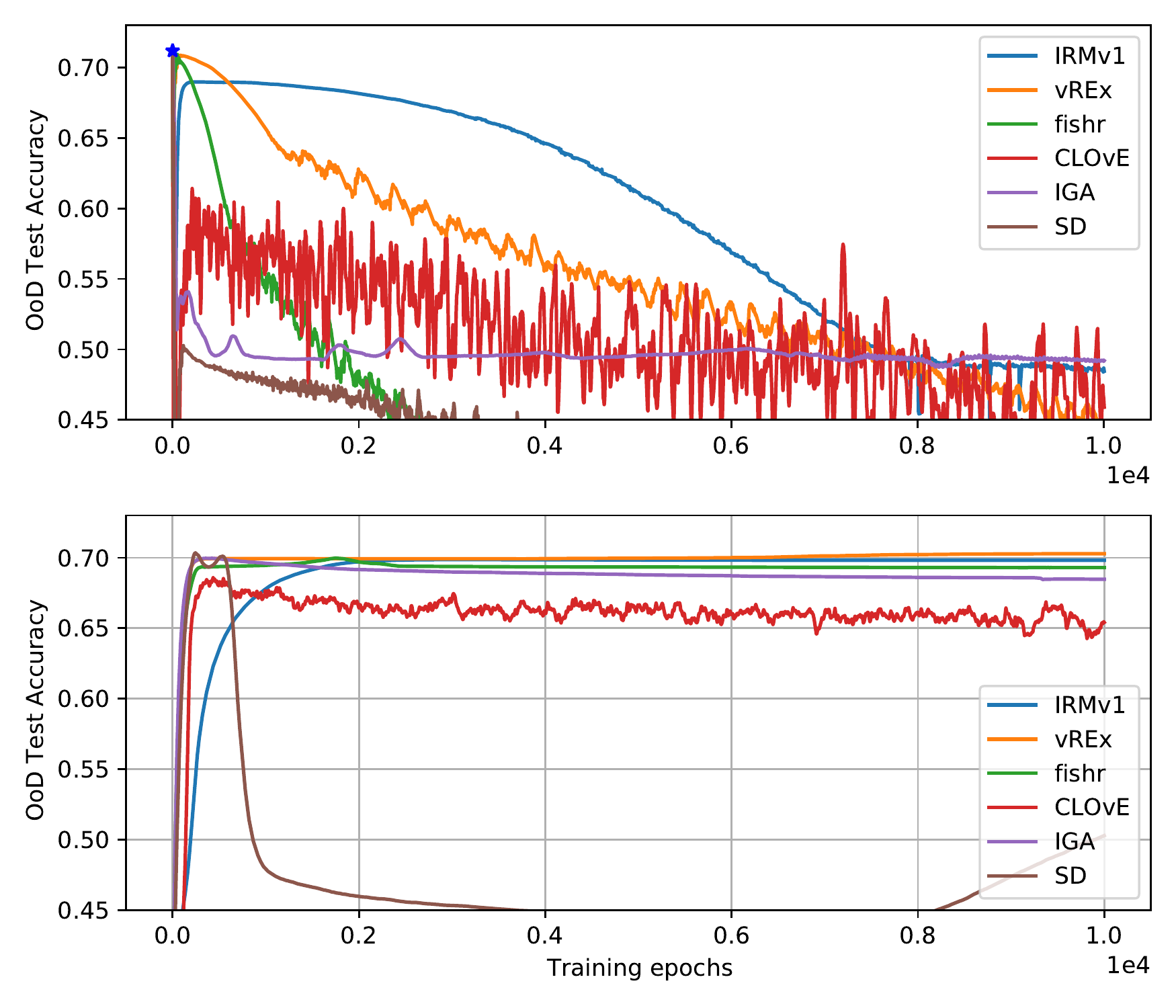}
    \caption{Test performance of OoD methods as a function of training epochs. Top: Six OoD methods are trained from a `perfect' initialization where only the robust feature is well learned. The blue star indicates the initial test accuracy. Bottom: The OoD methods are trained from the proposed (frozen) Bonsai representation. }
    \label{fig:generalization_difficulty}
\end{figure}

The top plot in figure \ref{fig:generalization_difficulty} shows how the OoD testing performance of six algorithms deviates from the performance of our perfect initialization. This might happen because
the chosen constraints have spurious solutions \citep{Kamath2021} or because the penalty terms are too weak to enforce the target constraints. Instead, the training process pulls the perfect initialization in the direction of the spurious feature (the color) which happens to be more predictive on the training data.

\section{Rich Feature Construction}
\label{sec:rfc}

This section presents tools for constructing rich representations. First, we describe a mathematically sound approach to the problem of constructing a rich set of diverse features and we introduce the notions of \emph{discovery} and \emph{synthesis} episodes. Then we show how to use Distributionally Robust Optimization (DRO) to cut on the synthesis episodes. Finally, we present the practical Bonsai algorithm that we use in Section~\ref{sec:experiments}.

\subsection{Feature discovery}

Intuitively, constructing additional features is desirable when using these features increases the system performance on a pertinent subset of examples. Best would of course achieve a large performance increase on large subsets of examples.

For the purposes of this section, let $\Phi_k(x)$ be a large vector containing all previously constructed $k$ features for pattern $x$. Our first step consists of defining an ensemble $P=\{D^1\dots{D^i}\dots\}$ of pertinent subsets $D^i$ of the training set. An effective way to choose a good ensemble of subsets is discussed at the end of this section. Having defined such subsets, we can define costs $C_i(\Phi,w)$ that measure
the quality of a feature set $\Phi$ measured on subset $D^i$:
\[
   C_i(\Phi,w) = \frac{1}{|D^i|} \sum_{(x,y)\in D^i} \ell(y, w^\top \Phi(x))
\]
where $w$ represent the weights of a linear layer and $\ell(y,\hat{y})$ is a convex loss. In the context of deep learning, considering a linear layer operating a large feature vector is not an unreasonable way to investigate the effectiveness of a representation \citep{ntk}. We can reweigh the training data in a manner that emphasizes the weaknesses of our current set of features, that is,
\begin{equation}
    \label{eq:r-rw}
    R_{rw} = \max_\lambda \min_w \sum_i \lambda_i C_i(\Phi_k,w)
\end{equation}
where the $\lambda_i$ coefficients are positive and sum to 1.  Let $\lambda^*_i$ be the pessimal mixture coefficients resulting from
optimization problem~\eqref{eq:r-rw}.  We can then learn a new set of features that help performance on this pessimal mixture,
\begin{equation}
    \label{eq:rprime-rw}
    R'_{rw} = \min_{w,\Phi} \sum_i \lambda^*_i C_i(\Phi,w)~.
\end{equation}
The main difference is that we are now training the features, yielding a new feature vectors $\Phi^*(x)$. If $R'_{rw}$  \eqref{eq:rprime-rw} is smaller than $R_{rw}$ \eqref{eq:r-rw}, then we know that $\Phi^*$ contains \emph{new useful features that were not present} in $\Phi_k$. This is the \emph{discovery phase}.

The next step consists in forming new feature vectors $\Phi_{k+1}(x)$ that contain the features present in both $\Phi_k$ and $\Phi^*$, a \emph{synthesis phase}. We can then iterate and obtain additional useful and diverse features at each iteration. The synthesis phase can be as simple as a vector concatenation. In the context of deep learning, however, one often has to use distillation, as discussed later in section~\ref{sec:rfc}.

The selection of a pertinent ensemble of subsets certainly affects which new features will be constructed at each iteration. In particular, it is desirable to make $R_{rw}$ as high as possible using a minimal number of subsets. This goal can be easily achieved by forming subsets containing examples that were either correctly classified or misclassified by the learning systems constructed by problem \eqref{eq:rprime-rw}.

\subsection{Using DRO}

We now show how a DRO reformulation of this process can cut the intermediate synthesis phase. Because the $C_i$ are convex in $w$,
we can first apply von Neumann's minimax theorem \citep[theorem~3]{simons1995} to problem \eqref{eq:r-rw} and obtain a DRO problem \citep{Ben-TalGN09}:
\begin{multline}
\label{eq:r-dro}
 R_{rw} ~=~ \max_\lambda \min_w \sum_i \lambda_i C_i(\Phi_k,w) \\
        = \min_w \max_\lambda \sum_i \lambda_i C_i(\Phi_k,w) \\
        = \min_w \max_i C_i(\Phi_k,w) ~=~  R_{dro}~. 
\end{multline}
The next step is to run this same DRO optimization while also learning the features
\begin{equation}
    \label{eq:rprime-dro}
    R'_{dro} = \min_{w,\Phi} \max_i C_i(\Phi,w)~.
\end{equation}
To understand how quantity $R'_{dro}$ relates to $R'_{rw}$, we can use the max-min inequality as follows:
\begin{multline*}
  R'_{dro} = \min_{w,\Phi} \max_\lambda \sum_i\lambda_i C_i(\Phi,w) \\
    \ge \max_\lambda \min_{w,\Phi} \sum_i \lambda_i C_i(\Phi,w) \\
    \ge \min_{w,\Phi} \sum_i \lambda^*_i C_i(\Phi,w) = R'_{rw}~.
\end{multline*}
In other words, if $R'_{dro}$ is smaller than $R_{dro}$, then $R'_{rw}$ is smaller than $R_{rw}=R_{dro}$, and the new feature vector $\Phi$ contains new and useful features. The advantage of this approach is that
problem \eqref{eq:rprime-dro} does not involve mixture coefficients $\lambda^*$. Therefore there is no need to solve \eqref{eq:r-dro} or \eqref{eq:r-rw}, and no need for a synthesis phase at each iteration. The synthesis phase is only needed to construct the final rich representation after the last iteration.

\subsection{The practical Bonsai algorithm}

We now describe a practical algorithm that implements the ideas discussed in the previous subsection in a manner that is usable with ordinary deep networks. 
The workhorse of this algorithm is the Robust Empirical Risk Minimisation (RERM) algorithm (Algorithm~\ref{alg:rerm}) which takes an ensemble of datasets $D^k$ representing multiple distributions and seeks neural network weights that simultaneously yields small errors for all these distributions. RERM is in fact a minimal form of DRO with overfitting control by cross-validation.

\begin{algorithm}[ht]
\caption{Robust Empirical Risk Minimization (RERM)}
\label{alg:rerm}
\begin{algorithmic}[1]
\STATE  \textbf{Required:} datasets $D^k = \{(x^k_i, y^k_i)\}_{i=1}^{n^k}$, for $k = 1, \ldots, N$; model $f$; learning rate $\alpha$
\STATE  Randomly initialize $f$
\WHILE[ \texttt{  // By validation}]{no overfit}
\STATE  Train on  datasets $D^1,\ldots,D^N$ by DRO: $f \leftarrow f - \alpha \cdot \nabla_f \left[\max_k \left(\frac{1}{|D^k|} \sum_{(x^k_i, y^k_i) \in D^k} \ell(f(x^k_i), y^k_i)\right)_{k=1}^N\right]$
\ENDWHILE
\STATE \textbf{return} $f$
\end{algorithmic}
\end{algorithm}

The Bonsai algorithm (Algorithm~\ref{alg:rich_feature}) first performs a predefined number of \emph{discovery episodes}, using RERM to repeatedly solve an analogue of problem \eqref{eq:rprime-dro} that constructs a model $f_k$ at each iteration, using an ensemble of subsets formed by selecting which examples were correctly or incorrectly recognized by the models $f_0\dots f_{k-1}$ constructed during the previous iterations.

The Bonsai algorithm performs a distillation-based \emph{synthesis episode}. The goal is to learn a representation network $\Phi(x)$ such that we can emulate the functions $f_k$ using a simple network with weights $w_k$ on top of $\Phi(x)$. To that effect, we use the $f_k$ models to compute pseudo-labels $y^k(x)$ for each example $x$. We then train a composite model with parameters $\Phi$, $w_1$, \dots, $w_K$ whose $k$ outputs are trained to replicate the pseudo-labels.

\par\noindent
\emph{Why use linear classifiers in synthesis episode (line 11)?}
\par
The goal is to perform the synthesis step by distillation into a network whose architecture is as close as possible as the architecture of the “source” networks. However the distillation network needs one head for each source network. The least intrusive way to implement multiple heads is to duplicate the very last layer, hence linear. The opposite approach would be to claim the whole network is a classifier and the feature extractor is the identity. In this case, we can get a perfect synthesis loss (Alg \ref{alg:rich_feature}, line 14) with an identity feature extractor which is obviously useless. We leave the non-linear classifier in \emph{synthesis phase} as a future work. 

\par\noindent
\emph{What if the first RERM round achieves zero errors (line~3)?}
\par
The training set of the first RERM round is the union of the data associated with all OOD training environments. Since RERM avoids overfitting using a validation set (Alg \ref{alg:rerm}, line 3), a perfect accuracy on both the merged training and validation sets means that the features discovered in the first round are already invariant in all training environments and perfectly predictive (100\% accuracy). Therefore no further rounds are necessary since we already have a solution. This is in fact a degeneracy of the invariant training concept.

\begin{algorithm}[t]
\caption{Bonsai algorithm (RFC)}
\label{alg:rich_feature}
\begin{algorithmic}[1]
\STATE \textbf{Input:} dataset $D$; the number of discovering rounds $K$
\STATE \texttt{// Discovery episodes}
\STATE $f_1 \leftarrow \text{RERM}(\{D\})$ 
\STATE Split $D$ into groups $A_1,B_1$ according to $f_1$. ($A_1=$ examples correctly classified by $f_1$, $B_1 = D \backslash A_1$)   
\STATE Available groups $P = \{A_1,B_1\}$
\FOR{$k \in [2,\ldots, K]$}
\STATE $f_k \leftarrow \text{RERM}(P)$
\STATE Split $D$ into groups $A_k,B_k$ according to $f_k$
\STATE $P \leftarrow P \cup \{A_k, B_k\}$
\ENDFOR
\STATE \texttt{// Synthesis episode}
\STATE Pick a feature extractor function $\Phi$, and $K$ linear classifiers $\omega_1,...\omega_k$ at random
\STATE Create $K$ groups of pseudo-labels ${y}^k$ by applying each $f_k$ on $D$ 
\STATE $A = A_i \cap ... \cap A_K$
\STATE Update $\Phi, \omega$ such that each pseudo-label ${y}^k$ is well learned by the corresponding classifier $\omega_k$ and $\Phi$: 
$\sum_{k=1}^{K} \frac{1}{|{A}|}\sum_{(x_i,{y}_i^k)\in {A}}\ell(\omega_k  \circ \Phi(x_i), {y}^k_i) + \frac{1}{|D\backslash {A}|}\sum_{(x_i,{y}_i^k)\notin {A}} \ell(\omega_k  \circ \Phi(x_i), {y}^k_i)$
\STATE \textbf{return} $\Phi, \{\omega_k\}_{k=1}^K$

\end{algorithmic}
\end{algorithm}

\section{Experiments}
\label{sec:experiments}

This section presents experimental results that illustrate how the rich representations constructed with RFC can help the OoD performance and reduce the performance variance of OoD methods.  Subsection~\ref{sec:rfc_ood} extends the experiments of Section~\ref{sec:spurious_tour} with RFC constructed representations. Subsection~\ref{sec:2nd_easy_feature} compares RFC-initialized OoD methods with recent methods that aim for the second easiest-to-find representation. Subsection~\ref{sec:camelyon} reports results obtained on the \textsc{Camelyon17} dataset \citep{bandi2018detection} that is part of the WILDS benchmark suite \citep{wilds2021}. The final subsection provides additional empirical observation that casts light on the hyper-parameter tuning process and on the feature construction process itself.

\setlength{\tabcolsep}{1.7mm}
\begin{table}[ht]
    \centering
    \begin{tabular}{l|c|c|c|c}
\toprule
 & Rand & ERM & 
 Bonsai& Bonsai-cf \\
\midrule
IRMv1 & 54.0$\pm$2.4  & 68.9$\pm$1.1  &  66.5$\pm$1.5         &\textbf{69.9$\pm$0.6}  \\
vREx  & 53.1$\pm$2.0  & 69.3$\pm$0.7  & \textbf{70.3$\pm$0.4} & 69.9$\pm$0.4  \\
SD    & 49.8$\pm$0.6  & 65.5$\pm$1.1  & 69.8$\pm$0.6          &\textbf{70.4$\pm$0.4}  \\
IGA   & 51.8$\pm$2.1  & 50.7$\pm$4.2  & 69.4$\pm$0.7          &\textbf{70.0$\pm$0.8}  \\
fishr & 53.9$\pm$1.5  & 69.2$\pm$0.9  & \textbf{70.2$\pm$0.4} & 69.4$\pm$0.8  \\
CLOvE & 43.9$\pm$4.2  & 63.7$\pm$2.5  & 67.1$\pm$3.8          & \textbf{68.4$\pm$0.8} \\
\midrule
ERM &  27.3$\pm$0.4 & 27.3$\pm$0.4 & 43.4$\pm$2.8 & 35.6$\pm$1.2 \\
\midrule
oracle & \multicolumn{4}{c}{72.1 $\pm$ 0.2 } \\
\bottomrule
    \end{tabular}
    \caption{OoD testing accuracy achieved on the \textsc{ColorMNIST}. The first six rows of the table show the results achieved by six OoD methods using respectively random initialization (Rand), ERM initialization (ERM), Bonsai initialization (Bonsai). The last column, (Bonsai-cf), reports the performance achieved by running the OoD algorithm on top of the frozen Bonsai representations. The seventh row reports the results achieved using ERM under the same conditions. The last row reminds us of the oracle performance achieved by a network using data from which the spurious feature (color) has been removed.}
    \label{tab:rich_rep_col025}
\end{table}

\subsection{Bonsai initialization helps all methods}
\label{sec:rfc_ood}

All experiments reported in this section use the \textsc{ColorMNIST} task \cite{irm} which consists of predicting labels that indicate whether the class of a colored digit image is less than 5 or not. The target label is noisy and only matches the digit class with probability 0.75 (correlation coefficient 0.5). Two training sets are provided where a spurious feature, the color of the digit, correlates with the target label with respective probabilities 0.8 and 0.9 (correlation coefficients 0.6 and 0.8). However, in the OoD testing set, the digit color is negatively correlated with the label (correlation coefficient -0.8). This testing protocol hits algorithms that rely on the spurious color feature because it happens to be more predictive than the robust feature in both training environments.
 
We compare six OoD training methods (IRMv1, vREx, SD, IGA, Fishr, CLOvE) and ERM after four types of initialization: (a) a random initialization with the popular Xavier method \citep{pmlr-v9-glorot10a}, (b) random initialization followed by several epochs of ERM, and (c) initialization with Bonsai representations, and (d) initialization with Bonsai representations that are subsequently frozen: the training algorithm is not allowed to update them (\mbox{Bonsai-cf}). The ERM initialization essentially consists of switching off the penalty terms defined by the various OoD method. This is comparable to the delicate penalty annealing procedures that are used by most authors \cite{irm, vrex, sd, rame2021fishr}. The Bonsai initialization was computed by two discovery phase iterations.

For all six OoD algorithms and four initialization strategies, we select one of five penalization weights, $\{10, 50, 100, 500, 1000\}$ for the SD method, $\{1000, 5000, 10000, 50000, 100000\}$ for the other methods. For ERM initialization, we also select among five numbers of pretraining epochs $\{50,100,150,200,250\}$. These hyper-parameters were selected by peeking at the test set performance.\footnote{The small size of the \textsc{ColoredMNIST} makes this hard to avoid. Tuning the hyper-parameters using the testing set favors in fact the ERM initialization because the test performance depends strongly on the number of pre-training epochs (Figure~\ref{fig:importance_of_rep}).} All experiments use the same 2-hidden-layers MLP network architecture (390 hidden neurons), Adam optimizer, learning rate=$0.0005$, $L_2$ weights regularization=$0.0011$ and binary cross-entropy objective function as the \textsc{ColoredMNIST} benchmark \cite{irm}. Further details are provided in Appendix~\ref{apdix:train_details_colorminst}.

Table~\ref{tab:rich_rep_col025} reports the OoD testing accuracies obtained under these conditions.  Bonsai initialization helps the OoD performance of most algorithms. Interestingly, the best results are achieved by freezing the Bonsai representation, which is consistent with the results of Section~\ref{sec:generalization_difficulty} showing that the OoD algorithm penalties are in fact insufficient to maintain the desired invariance constraints, even when initialized with the oracle weights. The bottom plot in Figure~\ref{fig:generalization_difficulty} shows that Bonsai initialization helps most OoD methods in this scenario as well, with the exception of the SD algorithm which penalizes the $L_2$ norm of logits in a manner that drives away the network from the oracle weights. Freezing the Bonsai representation doesn't prevent this from happening.

\subsection{Aiming for the second easiest-to-find feature}
\label{sec:2nd_easy_feature}

Recent work \cite{liu2021just,lff, Bao2021, ahmed2020systematic, creager2021environment} claims to achieve OoD generalization by discovering and using only the second easiest-to-find features. Although this strategy often works on datasets that were constructed to showcase OoD problems, the assumption that the second easiest features are the robust ones is unreasonable.

To illustrate this claim we construct a variant of the \textsc{ColoredMNIST} dataset by changing the noise levels to make the robust feature (the digit shapes) more predictive than the spurious features (the digit color). 

Table \ref{tab:pi_rfc} compares the six OoD methods using the frozen Bonsai representation on both \textsc{ColoredMNIST} and \textsc{InverseColoredMNIST}. All six methods achieve very comparable OoD testing accuracies. The ERM method fails on \textsc{ColoredMNIST} but performs quite well on \textsc{InverseColoredMNIST} because relying on the most predictive features is a good strategy for this task. In contrast, the algorithm PI \citep{Bao2021}, which aims for the second easiest features, performs well on \textsc{ColoredMNIST} but far worse on the easier \textsc{InverseColoredMNIST} task.

\begin{table}[t]
    \centering
    \begin{tabular}{l|c|c}
    \toprule
    Methods &\makecell{\textsc{ColoredMNIST}} & \makecell{\textsc{Inverse} \\ \textsc{ColoredMNIST}} \\
    \midrule
        IRMv1 & 69.9$\pm$0.6 &  80.3$\pm$2.2 \\
        vREx  & 69.9$\pm$0.4 &  84.0$\pm$1.2\\
        SD    & 70.4$\pm$0.4 &  81.9$\pm$1.3\\
        IGA   & 70.0$\pm$0.8 &  78.5$\pm$3.6 \\
        fishr & 69.4$\pm$0.8 &  82.6$\pm$1.4 \\ 
        CLOvE & 68.4$\pm$0.8 &  71.9$\pm$0.7\\
        ERM   & 35.6$\pm$1.2 &  71.7$\pm$0.7 \\
        \midrule
         PI      &  70.9$\pm$0.3  &51.0$\pm$4.7 \\
    \bottomrule
    \end{tabular}
    \caption{OoD test accuracy of PI and OoD/ERM methods on \textsc{ColoredMNIST} and \textsc{InverseColoredMNIST}. The OoD/ERM methods use a frozen Bonsai representation (Bonsai-cf).}
    
    \label{tab:pi_rfc}
\end{table}

\subsection{Bonsai initialization on a real-world task}
\label{sec:camelyon}

The \textsc{Camelyon17} dataset \citep{bandi2018detection} contains histopathological images accompanied by a label indicating whether the central region of the image contains a tumor. The images were collected from five different hospitals with potentially different imaging hardware and different procedures. The WILDS benchmark \citep{wilds2021} contains a task that uses this dataset with a very clear specification of which three hospitals are to be used as training data (302,436 images), which hospital is to be used for OoD generalization testing (85,054 images). The task also specifies multiple runs with different seeds in order to observe the result variability. Finally the task defines two ways to perform hyper-parameter selection: "IID Tune" selects hyper-parameters based on model performance on 33,560 images held out from the training data, "OoD Tune" selects hyper-parameters on the model performance observed on the fifth hospital (34,904 images).

We compare four different training methods, ERM, IRMv1, vREx, and CLOvE, using both ERM and Bonsai initialization with either two rounds (2-Bonsai) or three rounds (3-Bonsai) during the discovery phase. We also compare the effect of letting the training method tune the representation or freezing the representations obtained by the initialization procedure (ERM-cf, 2-Bonsai-cf, and 3-Bonsai-cf).

We strictly follow these procedures as well as the experimental settings suggested in the WILDS task. The network is a DenseNet121 model \citep{huang2017densely} trained by optimizing a cross-entropy loss with $L_2$ weight decay=$0.01$ using SGD with learning rate=$0.001$, momentum=$0.9$ and batch size=$32$. The penalty weights are selected from $\{0.5, 1, 5, 10, 50, 100, 500, 1000\}$ for IRMv1 and vREx, $\{0.0005, 0.001,0.005,0.01,0.05,0.1,0.5,1\}$ for CLOvE. The number of ERM pre-training iterations is selected in set from $\{0, 500, 1000, 5000, 10000\}$. Further details are provided in Appendix~\ref{apdix:camelyon17}.

Table \ref{tab:camelyon_full_results} reports the OoD testing accuracies obtained by using both the IID and OoD hyper-parameter tuning approach. Accuracies were averaged over five repetitions with different random seeds. The first block of rows reports accuracies obtained by all four methods using ERM initialization. These accuracies come with large error bars because they considerably vary across repetitions. As a consequence, the accuracies differences observed in this block are not significant. The second block of rows shows that freezing the representations does not significantly improve this situation. In contrast, using a Bonsai representation with two discovery rounds (2-Bonsai) consistently improves the accuracies obtained by all four methods using either the IID or OoD tuning approaches (third block of rows). Freezing the Bonsai representation provides an additional boost (fourth block of rows).

\citet{rosenfeld2022domain} claimed ERM may already discover enough features in the representation for OoD generalization. The second block in Table \ref{tab:camelyon_full_results} shows the ERM learned representation is not rich enough in the \textsc{camelyon17} case.

\begin{figure*}[t]
    \centering
    \includegraphics[width=\textwidth]{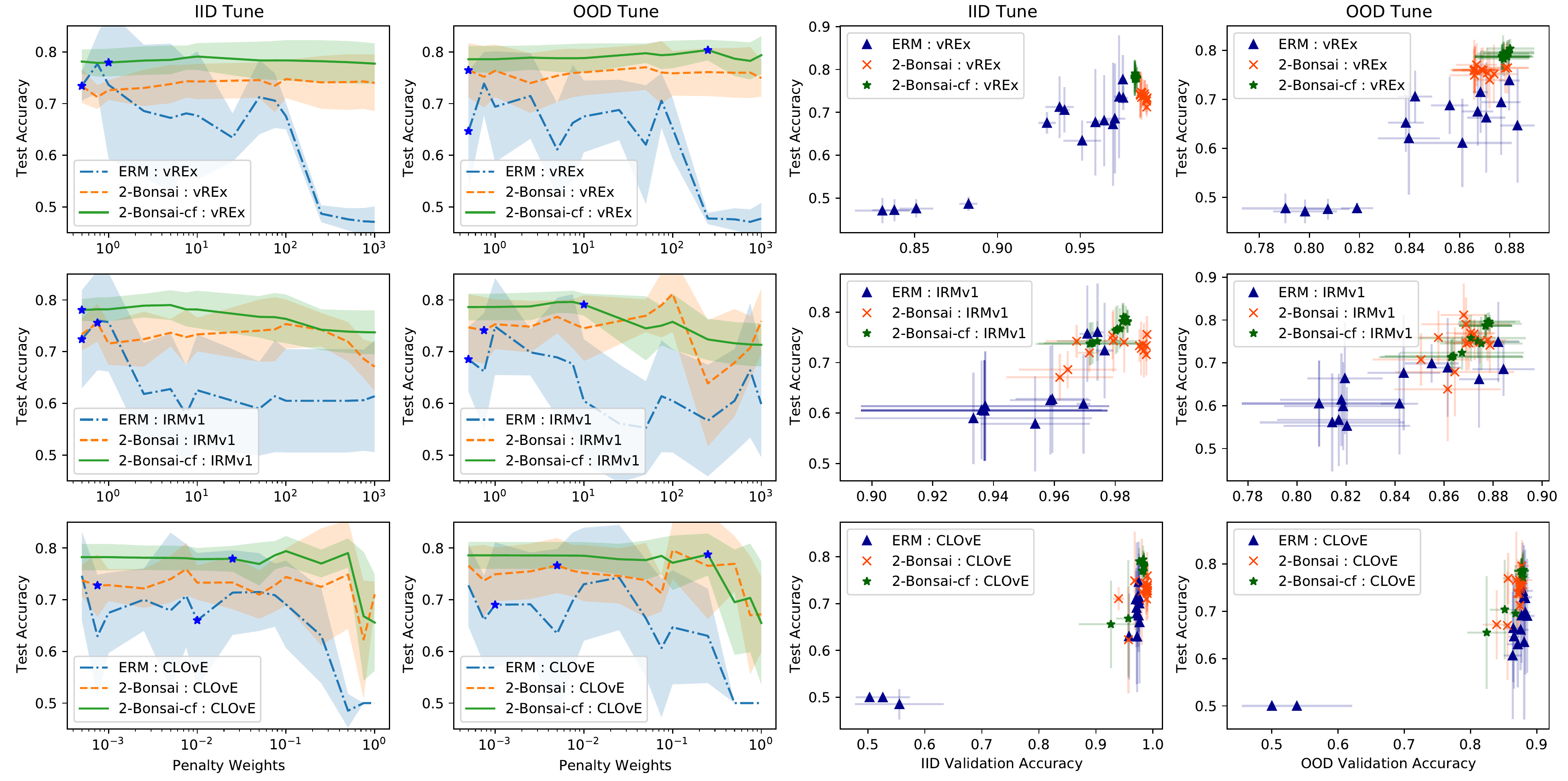}
    \caption{\textbf{Left half}: OoD testing accuracy as a function of the penalty weight. The six plots correspond to the IRMv1, vREx, and CLOvE algorithms with all other hyper-parameters selected using either the IID or OoD tuning method. Bonsai initialization makes these curves far more predictable than ERM initialization. Starts indicate the final penalty weight choice.~
    \textbf{Right half}: OoD testing accuracy as a function of the validation accuracy. Bonsai initialization reduces the variance of both the IID and OoD validation performances, making them far more reliable indicators of the actual OoD testing performance.}
    \label{fig:lambda_valid_test}
\end{figure*}

\begin{table}[t]
    \centering
    
    \begin{tabular}{l|l|c|c}
        \toprule
Network        & Methods & \multicolumn{2}{c}{Test Acc} \\
Initialization &         & IID Tune & OoD Tune \\
\midrule
$\times$ & ERM & 66.6$\pm$9.8 & 70.2$\pm$8.7 \\
ERM & IRMv1 & 68.6$\pm$6.8 & 68.5$\pm$6.2\\
ERM & vREx & 69.1$\pm$8.1 & 69.1$\pm$13.2\\
ERM & CLOvE & 71.7$\pm$10.2 & 69.0$\pm$12.1 \\
\midrule
ERM-cf &ERM & $\times$ & $\times$ \\
ERM-cf &IRMv1 & 69.6$\pm$10.5 & 70.7$\pm$10.0  \\
ERM-cf &vREx & 69.6$\pm$10.5 & 70.6$\pm$10.0   \\
ERM-cf &CLOvE & 69.6$\pm$10.5 & 69.2$\pm$9.5   \\
 \midrule
2-Bonsai & ERM   & 72.8$\pm$3.2   & 74.7$\pm$4.3 \\
2-Bonsai & IRMv1 & 71.6$\pm$4.2 & 75.3$\pm$4.8  \\
2-Bonsai & vREx  &  73.4$\pm$3.3 & 76.4$\pm$5.3  \\
2-Bonsai & CLOvE & 74.0$\pm$4.6 &76.6$\pm$5.3    \\
\midrule
2-Bonsai-cf & ERM   & 78.2$\pm$2.6 & 78.6$\pm$2.6\\
2-Bonsai-cf & IRMv1 & 78.0$\pm$2.1 & 79.1$\pm$2.1 \\
2-Bonsai-cf & vREx  & 77.9$\pm$2.7 & 79.5$\pm$2.7 \\
2-Bonsai-cf & CLOvE & 77.8$\pm$2.2 & 78.6$\pm$2.6 \\
\midrule
3-Bonsai-cf & ERM   & 72.9$\pm$5.3 & 73.3$\pm$5.3 \\
3-Bonsai-cf & IRMv1 & 72.7$\pm$5.5 & 75.5$\pm$3.8 \\
3-Bonsai-cf & vREx  & 72.7$\pm$5.4 & 75.1$\pm$5.3 \\
3-Bonsai-cf & CLOvE  & 72.8$\pm$5.4 & 73.2$\pm$7.1 \\
         \bottomrule
    \end{tabular}
    \caption{Test Accuracy on the \textsc{camelyon17} dataset. The hyper-parameter tuning process is performed on either the iid validation or the OoD validation set (``IID/OoD Tune''). We test ERM pretrained initialization, 2-rounds, and 3-rounds Bonsai representation. As to the learning methods, we test ERM, IRMv1, vREx, and CLOvE. When freezing the representation and training the top-layer classifier only, we get ``-cf'' methods. The standard deviation is calculated on 5 random seeds [0-4].}
    \label{tab:camelyon_full_results}
\end{table}

Using a Bonsai representation with three discovery rounds (3-Bonsai-cf) does not work as well. In fact, the features extracted during the third discovery phase round are not as predictive as the first two rounds (Table~\ref{tab:camelyon17_rfc_performance}). More discovery rounds also increase the difficulty of the synthesis phase, as we want to distillate more features (including poor ones) into the same fixed-size representation.

Much to our surprise, Bonsai initialization consistently boosts the accuracies of both the ERM and OoD methods, using either the IID or OoD tuning method. The frozen Bonsai representations can even help ERM outperform earlier comparable results reported on the WILDS leaderboard\footnote{\url{https://wilds.stanford.edu/leaderboard}} by about 5\%.

\subsection{Further observations}

\subsubsection{Hyper-parameter tuning}
\label{sec:hyper-paraemeter-model-selection-difficulty}

Figure \ref{fig:importance_of_rep} and \ref{fig:generalization_difficulty} illustrate how the OoD generalization performance of many OoD methods depends strongly on hyper-parameters such as the number of pretraining epochs, the penalty weights, the learning epochs. This is in fact a consequence of the optimization-generalization dilemma itself. It is simply difficult to simultaneously ensure good OoD generalization performance and run a stable and efficient optimization process.

The left half of Figure~\ref{fig:lambda_valid_test} shows OoD testing accuracies for the \textsc{Camelyon17} task as a function of the penalty weights, with all other hyper-parameters chosen using either the IID or OoD tuning method. With ERM pretraining, the OoD testing performance of all three OoD methods (IRMv1, vREx, CLOvE) depends very chaotically on the penalty weight. In contrast, with a frozen Bonsai representation, the OoD testing performance of OoD methods, as a function of the penalty weight, follows a much smoother curve.

The right half of Figure~\ref{fig:lambda_valid_test} shows the relation between the IID/OoD validation accuracies and the OoD testing accuracies for three OoD methods using both ERM and Bonsai initialization. Bonsai initialization reduces the variance of both the IID and OoD validation performances, making them far more reliable indicators of the actual OoD testing performance.

\subsubsection{The value of the synthesis episode}
\label{sec:discovery_phase_performance_camelyon17}

The \textsc{ColoredMNIST} and \textsc{InverseColoredMNIST} experiments (Table~\ref{tab:pi_rfc}) show that the robust feature can be discovered during different rounds of the discovery phase. We can therefore wonder whether the discovery phase already produces the correct invariant representation during one of its successive rounds.

This is not the case in general. Table~\ref{tab:camelyon17_rfc_performance} reports the OoD testing accuracies of the classifiers constructed during the first three rounds of the discovery phase. All three accuracies are substantially worse than the accuracies achieved by any algorithm using a frozen 2-Bonsai-cf representation (Table~\ref{tab:camelyon_full_results}). This indicates that these higher accuracies are obtained by simultaneously exploiting features discovered by different rounds of the discovery phase. Making them all simultaneously available is indeed the role of the synthesis phase.

 \setlength{\tabcolsep}{6.5mm}
\begin{table}[ht]
    \centering
    \begin{tabular}{c|c|c}
    \toprule
    Round 1 & Round 2 & Round 3 \\
     \midrule
66.6$\pm$9.8 & 73.2$\pm$5.7 & 61.8$\pm$10.2 \\
\bottomrule
    \end{tabular}
    \caption{OoD test accuracies for the models constructed by the first three discovery rounds for the \textsc{Camelyon17} task. The first round amounts to performing ERM. The second round extracts a useful set of features. The third round extracts comparatively weaker features. All these accuracies remain substantially worse than those achieved by training a system on top of the combined representation computed during the synthesis phase (see Table~\ref{tab:camelyon_full_results}).}
    \label{tab:camelyon17_rfc_performance}
\end{table}
\section{Conclusion}
\label{sec:discussion}

This work makes several contributions:
\begin{itemize*}
\item We point out the severity of the optimization-generalization dilemma in the OoD setup, showing that the various penalties introduced by OoD methods are either too strong to optimize or too weak to achieve their goals.
\item We propose to work around the problem by seeding the networks with a rich representation that contains a diversity of features readily exploitable by the algorithm. We formalize this objective, and we describe an algorithm, the Bonsai algorithm, that constructs such rich representations.
\item We show that Bonsai initialization helps a variety of OoD methods achieve a better OoD testing performance. Interestingly, when we additionally prevent the learning algorithm from modifying the Bonsai representations, we not only observe a further boost in the performance of OoD methods, but we also raise the performance of ERM to the same level, substantially outperforming previous comparable results on the \textsc{Camelyon17} dataset for examples.
\item
Finally, we also show that Bonsai initialization facilitates both IID and OoD variants of hyper-parameter tuning and model selection.
\end{itemize*}
Therefore, it appears that the inductive bias that comes with a broad set of diverse features brings considerable benefits to the various invariant/OoD training methods proposed in the recent literature.
\section{Acknowledgements}
The authors acknowledge support from the National Science Foundation (NSF Award 1922658) and from the Canadian Institute for Advanced Research (CIFAR).



\bibliography{references}
\bibliographystyle{icml2022}

\onecolumn
\vfill




\clearpage
\onecolumn
\appendix

\section{MAML-IRM resembles vREx+Fish}
\label{apdix:mamlirm=vrex+gm}

We omit the MAML-IRM method in our experiments because we can show that minimizing its cost amounts to minimizing a mixture of the vREx and Fish costs.

Notations: 
\begin{itemize*}
    \item $\mathcal{E}$: indicates a set of environments.
\item $\theta$: indicates the model parameters.
\item $L_i(\theta)$: indicates an ERM loss (e.g. MSE, cross-entropy) of a model parameterized by $\theta$ on environments $i$.
\item $\bar{g_i} = L_i^{'}(\theta)$: is the gradients of $L_i(\theta)$.
\item $\bar{H_i} = L_i^{''}(\theta)$: is the Hessian of $L_i(\theta)$.
\end{itemize*}

Let $U_i(\theta) = \theta - \alpha L_i^{'}(\theta)$ denote the updated parameters after performing a SGD iteration on environments $i$.
The MAML-IRM loss can be expressed as:
\begin{align}
    L_\text{maml-irm} = \mathbb{E}_s [L_j(U_i(\theta))] + \lambda \sqrt{Var_s[L_j(U_i(\theta))]}
\end{align}
where the notation $\mathbb{E}_s$ and $Var_s$ respectively denote the average and the variance with respect to all pairs of distinct environment $s = \{i,j | i\in \mathcal{E}, j \in \mathcal{E}, i \neq j\}$, and where $\lambda$ is a hyper-parameter.

According to~\cite{nichol2018first}, the gradients of the first term is:
\begin{align}
    \frac{\partial (\mathbb{E}_s [L_j(U_i(\theta)))]}{ \partial \theta} &= \mathbb{E}_s [\bar{g_j} - 2\alpha \bar{H_i}\bar{g_j}] + O(\alpha^2)
\end{align}
Note that $\mathbb{E}_s [-2\bar{H_i}\bar{g_j}] = \mathbb{E}_s [-\frac{\partial\left\langle  g_i, g_j \right\rangle}{\partial \theta}]$ is in fact the gradients of $-\left\langle  g_i, g_j \right\rangle$, the Fish penalty.  

We now turn out attention to the second term $\sqrt{Var_s[L_j(U_i(\theta))]}$. Expanding $L_j(U_i(\theta))$ with a Taylor series gives:
\begin{align}
    L_j(U_i(\theta)) &= L_j(\theta) + \left\langle L_j^{'}(\theta),(U_i(\theta) - \theta)) \right\rangle + O(\alpha^2) \\
    &= L_j(\theta) - \alpha \left\langle L_i^{'}(\theta),L_j^{'}(\theta)\right\rangle + O(\alpha^2) \quad\quad\quad\quad (\longleftarrow U_i(\theta) = \theta - \alpha L_i^{'}(\theta)) \\
    &=  L_j(\theta) - \alpha\left\langle \bar{g_i},\bar{g_j}\right\rangle + O(\alpha^2)
\end{align}
Therefore
\begin{align}
    Var_s(L_j(U_i(\theta)))&= Var_s[L_j(\theta) - \alpha \left\langle \bar{g_i},\bar{g_j}\right\rangle]  + O(\alpha^2) \\
    &=Var_s[L_j(\theta)] + \alpha^2 Var_s[\left\langle \bar{g_i},\bar{g_j}\right\rangle] - 2\alpha \text{Cov}_s[L_j(\theta), \left\langle \bar{g_i},\bar{g_j}\right\rangle] + O(\alpha^2)  \notag \\
    &= Var_s[L_j(\theta)] - 2\alpha \text{Cov}_s[L_j(\theta), \left\langle \bar{g_i},\bar{g_j}\right\rangle] + O(\alpha^2)  \notag\\
    &=  Var_s[L_j(\theta)] - 2\alpha \{ \mathbb{E}_s[L_j(\theta)\left\langle \bar{g_i},\bar{g_j}\right\rangle] - \mathbb{E}_s [L_j(\theta)]\mathbb{E}_s[\left\langle \bar{g_i},\bar{g_j}\right\rangle]\} + O(\alpha^2) \notag  \notag \\
    &=  Var_s[L_j(\theta)] - 2\alpha  \mathbb{E}_s[(\frac{L_i(\theta)+L_j(\theta)}{2}-\mathbb{E}[L(\theta)])\left\langle \bar{g_i},\bar{g_j}\right\rangle] + O(\alpha^2)  \notag
\end{align}
The first term of this expression, $Var_s[L_j(\theta)]$, penalizes a high variance of the loss across environments. It is equal to the vREx penalty. The second term, $- 2\alpha \mathbb{E}_s[(\frac{L_i(\theta)+L_j(\theta)}{2}-\mathbb{E}[L(\theta)])\left\langle\bar{g_i},\bar{g_j}\right\rangle]$ is a weighted average of $\left\langle g_i,g_j\right\rangle$, that is a smoothed Fish penalty. 

In conclusion, optimizing the MAML-IRM cost amounts to optimizing a $\lambda$ controlled mixture of the vREx and Fish costs.

\section{GroupDRO interpolates environments while vREx extrapolates.}
\label{apdix:groupdro_vrex_inter_extra}
The vREx objective function can be expressed as: 
\begin{align}
    L_\text{vrex} = \mathbb{E}_{e\in\mathcal{E}} (L_e) + \lambda Var_{e\in\mathcal{E}}(L_e)
\end{align}

The GroupDRO objective function is a mixture of the per-environment costs $L_e$ with positive coefficients:
\begin{align}
    L_\text{groupDRO} = \mathbb{E}_{e\in\mathcal{E}} (p_e L_e)
\end{align}
where the adjustable mixture coefficients $p_e\geq0$, $\sum_{e\in\mathcal{E}}p_e = 1$, are treated as constaants for computing  the gradients  $\frac{\partial L_e}{\theta}$.  

The gradient of these two cost functions are:
\begin{align}
    \frac{\partial L_\text{vrex}}{\theta} &= \mathbb{E}_{e\in\mathcal{E}}([2\lambda(L_e - \mathbb{E}_iL_i) + 1]g_e) \\
    \frac{\partial L_\text{groupDRO}}{\theta} &= \mathbb{E}_{e\in\mathcal{E}}(p_eg_e)
\end{align}
where $g_e= \frac{\partial L_e}{\theta}$ is the gradients of network weights $\theta$ on environment $e$. 

Because the $p_e$ mixture coefficients are always positive, it is easy to see that GroupDRO follows a direction aligned with a convex combination of the per-environment gradients. In contrast, vREx can follow 
a direction that is outside this convex hull because the coefficients $\mathbb{E}_iL_i) + 1]$ can be positive or negative). 



\section{Loss landscape of OoD methods}
\label{apdix:lowdim_losslandscape}

Here we visualize the loss landscape of some of OoD penalties on a synthetic two-dimensional problem, TwoBits, which was introduced by \citep{Kamath2021} as a simplified version of the \textsc{coloredMNIST}. 
TwoBits is a binary classification problem $Y=\pm1$ with two binary inputs $X_1=\pm1$ and $X_2=\pm1$ distributed as follows:
\begin{align*}
    Y ~ &\sim ~ \text{Rademacher}(0.5) \\
    X_1 &\sim ~ Y \cdot \text{Rademacher}(\alpha_e) \\
    X_2 &\sim ~ Y \cdot \text{Rademacher}(\beta_e)
\end{align*}
where $\text{Rademacher}(\alpha)$ denotes the law of a random variable taking value $-1$ with probability $\alpha$ and taking $+1$ probability $1{-}\alpha$.
The training algorithms observe two training environments, $(\alpha_e, \beta_e) \in \{(0.1, 0.1), (0.1, 0.3)\}$. The four input patterns $(X_1,X_2)$ are represented by four points $\{\Psi(1,1),\Psi(-1,-1),\Psi(1,-1),\Psi(-1,1)\}$ in the representation space where $\Psi$ can represent any network architectures with numerical outputs. Following \citep{Kamath2021}, we use a mean squared loss and focus on the symmetric case $\Psi(-x)=-\Psi(x)$. The representation space can therefore be displayed with only two dimensions, $\Psi(1,-1)=-\Psi(-1,1)$ and $\Psi(1,1)=-\Psi(-1,-1)$. 

\begin{figure}
    \centering
    \includegraphics[width=\textwidth]{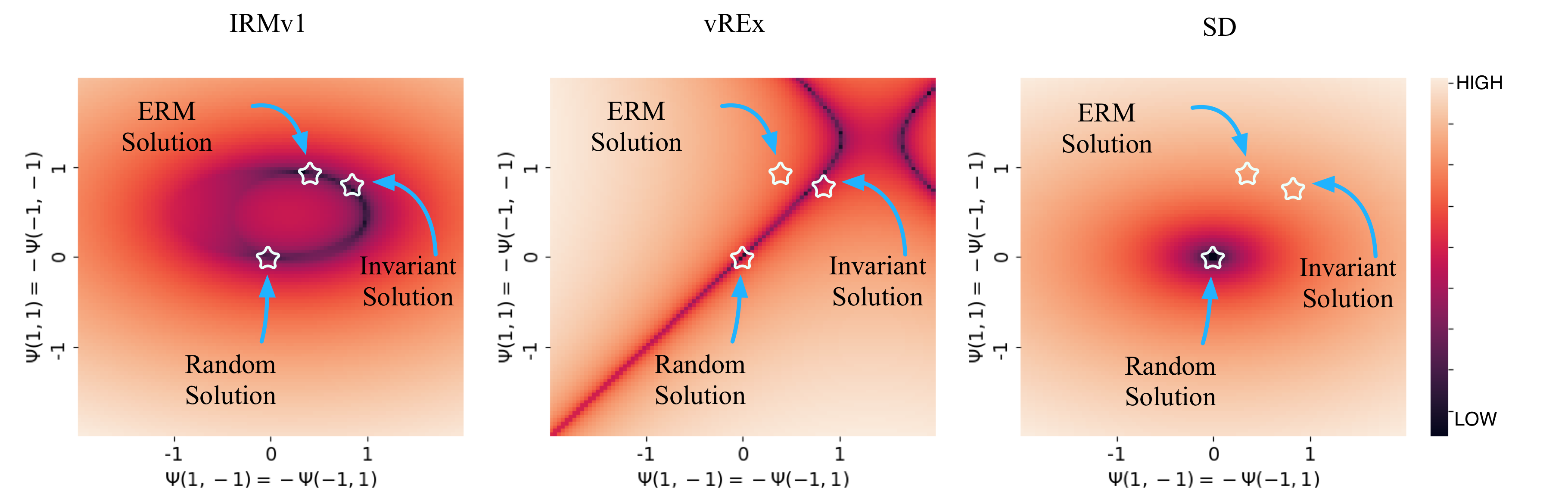}
    \caption{The IRMv1, vREx, and SD landscapes show a challenging non-convex landscape in the two-dimensional TwoBits problem. In particular, the path between the ERM solution and the invariant solution often involves climbing the loss landscape.}
    \label{fig:isolatedspace}
\end{figure}

Figure \ref{fig:isolatedspace} shows a heat map of the penalty terms of three OoD methods (IRMv1, vREx, SD) as a function of the chosen representation. The stars denote three solutions: (a) the Invariant solution which only uses feature $X_1$ because this is the feature whose correlation with the label remains the same across the training environments, (b) the ERM solution which uses both features, and (c) a random feature initialization with small variance for which the
representations $\Psi(1,1), \Psi(1,-1)$ are close to zero.  

All three OoD methods have low penalties when the $\Psi(1,1), \Psi(1,-1)$ are close to zero. This explains why random initialization performs so poorly with these methods. In contrast, pretraining with ERM leads to a new initialization point that is away from the origin and close to the ERM solution. The OoD performance then depends on the existence of a good optimization path between this initialization and the Invariant solution. Alas Figure~\ref{fig:isolatedspace} shows a lot of optimization difficulties such as finding a solution that lies at the bottom of an elongated ravine (ill-conditioning).  In conclusion, the impact of the number of ERM pretraining epochs is essentially unpredictable.  

\section{Experimental details for the ColoredMNIST experiments}
\label{apdix:train_details_colorminst}
We use the original \textsc{ColoredMNIST} dataset \cite{irm} with two training environments $(0.25, 0.1), (0.25, 0.2)$. The target label correlates with the invariant feature (the  digit shape) with a probability 0.75. The sirious feature (color) correlates with the target label with a probability 0.8 and 0.9, respectively. Each training environment contains $25000$ images where the size of each image is $2\times14\times14$. For all \textsc{ColoredMNIST} experiments, we use a fully connected neural network with 3 layers (392 (input dim) $\times390\times390\times1$), trained with the Adam optimizer with learning rate 0.0005. We use a L2 weights regularization with parameter 0.0001 for \textsc{InverseColoredMNIST} tasks and 0.0001 in the regular \textsc{ColoredMNIST} tasks. 
For the CLOvE method, we use a Laplacian kernel $k(r,r_0)=exp(\frac{-|r-r_0|}{0.4})$ \cite{kumar2018trainable} with mini-batch size 512. All other methods train using full batches. For the ERM baseline and for computing the oracle performance, we search the L2 regularization parameter in $\{0.0001, 0.0005, 0.001, 0.005, 0.01\}$. We run each experiment 10 times to get the standard deviation. 

\subsection{Hyper-parameter searching space}
Table \ref{tab:penalty-weights-searching-space} shows the penalty weights searching space for all OoD  methods in the \textsc{ColoredMNIST} experiments. Table \ref{tab:training_epoch_search_space} shows the training epochs searching space for different OoD  methods and network initialization/representation on the \textsc{ColoredMNIST} dataset.

\begin{table}[ht]
  \label{tab:penalty-weights-searching-space}
  \caption{Penalty weight search space for both the \textsc{ColoredMNIST} and \textsc{InverseColoredMNIST} datasets. }
  \label{sample-table}
  \centering
  \begin{tabular}{l|c|c}
    \toprule
                     &\textsc{ColoredMNIST}     & \textsc{InverseColoredMNIST}  \\
    \midrule
    IRMv1            & $10000\times\{0.1, 0.5, 1, 5, 10\}$& $10000\times\{0.1, 0.5, 1, 5, 10\}$ \\
    vREx             & $10000\times\{0.1, 0.5, 1, 5, 10\}$& $10000\times\{0.1, 0.5, 1, 5, 10\}$ \\
    IGA              & $10000\times\{0.1, 0.5, 1, 5, 10\}$& $10000\times\{0.1, 0.5, 1, 5, 10\}$ \\
    CLOvE            & $10000\times\{0.1, 0.5, 1, 5, 10\}$& $10\times\{0.1, 0.5, 1, 5, 10\}$ \\
    Fishr            & $10000\times\{0.1, 0.5, 1, 5, 10\}$&  $10000\times\{0.1, 0.5, 1, 5, 10\}$\\
     SD              & $100  \times\{0.1, 0.5, 1, 5, 10\}$& $\{0.05, 0.1, 0.5, 1,5\}$\\
    RSC              & $(0.995,0.98) \times \{0.95,0.97,0.98,0.99,1\}$& - \\
    LfF              & $\{0.1, 0.2, 0.3, 0.4, 0.5\}$& - \\
    Fish             & $0.001 \times\{0.1, 0.5, 1, 5, 10\}$ &- \\
    \bottomrule
  \end{tabular}
\end{table}

\begin{table}[ht]
    \centering
        \caption{The number of training epochs search space for the \textsc{ColoredMNIST} dataset, with $i \in [0,24]$.}
    \label{tab:training_epoch_search_space}

    \begin{tabular}{l|c|c|c}
    \toprule
            & Rand/ERM & Bonsai & Bonsai-cf\\
    \midrule
         IRMv1  &    $i\times 20$ & $i\times2$  &   $i\times 125$  \\
         vREx   &    $i\times 20$ & $i\times2$  &   $i\times 20$  \\
         IGA   &    $i\times 20$ & $i\times1$  &   $i\times 20$  \\
         CLOvE   &    $i\times 30$ & $i\times1$  &   $i\times 20$  \\
         Fishr   &    $i\times 20$ & $i\times1$  &   $i\times 20$  \\
         SD   &    $i\times 20$ & $i\times1$  &   $i\times 20$  \\
         RSC   &    $i\times 1$ &  -  &   -  \\
         LfF &$i\times 20$ & - & - \\
         Fish &$i\times 20$ & - & -\\
    \bottomrule
    
    \end{tabular}
\end{table}

\subsection{Bonsai algorithm}
For all \textsc{ColoredMNIST} experiments, we use a 2-rounds Bonsai \emph{discovery phase} trained with respectively 50 and 500 epochs.  Then we train 500 epochs for the distillation network of the Bonsai \emph{synthesis phase}. For the \textsc{InverseColoredMNIST} experiments, we again use a 2-rounds Bonsai \emph{discovery phase} trained with respectively 150 and 400 epochs. We choose these training epochs because they can maximize the IID validation performance during each round.

\subsection{PI training}
We use the original implementation from PI \cite{Bao2021}. Because the PI algorithm is closely related to the \emph{discovery phase}, we use the same hyper-parameters and settings.

\section{Experimental details for the \textsc{Camelyon17} experiments}
\label{apdix:camelyon17}

We strictly follow the implementation of the \textsc{Camelyon17} task in the WILDS benchmark \citep{wilds2021}. For the results presented in section \ref{sec:hyper-paraemeter-model-selection-difficulty}, we additionally search the penalty weights in the set $\{0.5,0.75, 1, 2.5, 5, 7.5, 10, 25, 50, 75, 100, 250, 500, 750, 1000\}$ for IRMv1 and vREx methods, and the set $\{0.5,0.75, 1, 2.5, 5, 7.5, 10, 25, 50, 75, 100, 250, 500, 750, 1000\}\times10^{-3}$. The CLOvE method require a kernel function, we choose the Laplacian kernel $k(r,r_0) = exp(\frac{-|r-r_0|}{l})$  \cite{kumar2018trainable} where $l$ is a positive scalar. For the CLOvE baseline with an ERM pretrained initialization (the fourth row of table \ref{tab:camelyon_full_results}), we test the scalar $l\in \{0.1, 0.2\}$ and choose the better one $l = 0.2$. For the other CLOvE experiments on \textsc{Camelyon17}, we choose $l=0.1$.

We train the \emph{synthesis phase} 20 epochs and the other methods/phase 10 epochs. Hyper-parameter tuning strictly follows the IID and OoD tuning process described in the WILDS task. We use a L2 weights regularization $1e-6$ during the \emph{synthesis phase} to help it get a lower training loss on the pseudo-labels. During any further training that updates the weights of the learned representation, we keep the L2 weights regularization to be the same as $1e-6$. Otherwise, a stronger L2 weights regularization will destroy the learned representation. We also tried other L2 regularization weights in $\{1e-2, 1e-4, 1e-6\}$. Table \ref{tab:synthesis_quality} shows the synthesis quality with different (\emph{synthesis phase}) L2 weights decay. Two smaller L2 weights decay hyper-parameters $\{1e-4, 1e-6\}$ can arrive at a good synthesis quality. The corresponding test performances on the frozen representation ``2-Bonsai-cf'' of the two smaller hyper-parameters are higher too (Table \ref{tab:more_synthesis_l2}). Table \ref{tab:more_synthesis_l2} shows that the "2-Bonsai-cf" representation can also reliability gain a high performance once the synthesis quality is good.

After the \emph{synthesis phase}, RFC provides us a rich representation $\Phi$ and $K$ linear classifiers $\omega_1,\dots, \omega_K$. In the downstream tasks, such as OoD/ERM training, we will keep the representation $\Phi$ and initialize the top-layer classifier $\omega$. There are at least two ways to initialize it: 1) initialize $\omega$ as the average of $\omega_1,\dots, \omega_K$ with the hope that the initial top-layer classifier uses all discovered features. 2) randomly initialize $\omega$. Table \ref{tab:rand_average} shows the test performance of OoD/ERM methods with each top-layer initialization method. None of the two top-layer initialization methods significantly outperforms the other one. We choose the first top-layer initialization method in all main experiments because of the interpretation.

\begin{table*}[ht]
    \centering
        \caption{Test accuracy of OoD methods (IRMv1, vREx) and ERM methods. Three \emph{synthesis phase} L2 weights decay $\{1e-2, 1e-4, 1e-6\}$ are tested. All the other settings are the same as the main results in Table \ref{tab:camelyon_full_results}. }
    \label{tab:more_synthesis_l2}
    \begin{tabular}{l|l|l|c|c}
        \toprule
        Synthesis phase & Network  & Methods & \multicolumn{2}{c}{Test Acc} \\
        L2 weights decay & Initialization  &   & IID Tune & OoD Tune \\
        \midrule
        $1e-6$ &    2-Bonsai-cf & ERM  & 78.2$\pm$2.6 & 78.6$\pm$2.6\\
        $1e-6$ &    2-Bonsai-cf & IRMv1  & 78.0$\pm$2.1 & 79.1$\pm$2.1 \\
        $1e-6$ &    2-Bonsai-cf & vREx  & 77.9$\pm$2.7 & 79.5$\pm$2.7 \\
        \midrule
        $1e-4$ & 2-Bonsai-cf & ERM    & 77.8$\pm$1.7 & 78.8$\pm$2.3 \\
        $1e-4$ & 2-Bonsai-cf & IRMv1    & 77.7$\pm$1.7 & 78.9$\pm$2.3 \\
        $1e-4$ & 2-Bonsai-cf & vREx   & 77.9$\pm$1.7 & 79.7$\pm$1.7 \\
        \midrule
        $1e-2$ & 2-Bonsai-cf & ERM   & 75.2$\pm$7.8 & 75.5$\pm$7.4   \\
        $1e-2$ & 2-Bonsai-cf & IRMv1    &75.0$\pm$7.9 & 75.4$\pm$7.5 \\
        $1e-2$ & 2-Bonsai-cf & vREx   & 75.4$\pm$7.7& 75.8$\pm$7.3  \\
        \bottomrule
    \end{tabular}
\end{table*}
\begin{table*}[ht]
    \centering
    \caption{The train and IID-validation performance of the \emph{synthesis phase}. Note that it uses the pseudo-labels instead of the true labels as $Y$. Three \emph{synthesis phase} L2 weights decay $\{1e-2, 1e-4, 1e-6\}$ are tested.  }
    \label{tab:synthesis_quality}
    \begin{tabular}{c|c|c}
            \toprule
           (Synthesis phase) L2 weights decay & Train accuracy & IID-validation accuracy\\
           \midrule
          $1e-6$ &99.7$\pm$0.0 & 97.4$\pm$0.3 \\
          $1e-4$ &99.6$\pm$0.1 & 97.4$\pm$0.2 \\
          $1e-2$ &93.9$\pm$0.7 & 94.9$\pm$0.5 \\
          \bottomrule
    \end{tabular}
\end{table*}

\setlength{\tabcolsep}{4mm}
\begin{table*}[ht]
    \centering
     \caption{Test performance of IRMv1, vREx, and ERM methods on a 2 rounds Bonsai representation. The top-layer classifier is initialized by either the average of $\omega_1,\dots \omega_K$ (Average) or a random initialization (Random). When freezing the representation and training the top-layer classifier only, we get the “-cf” methods.  }
    \begin{tabular}{c|c|c|c||c|c}
    \toprule
Network Initialization & Methods & \multicolumn{2}{c}{Average} & \multicolumn{2}{c}{Random} \\
        & & IID Tune & OOD Tune &  IID Tune & OOD Tune \\
        \midrule
2-Bonsai&ERM & 72.8$\pm$3.2   & 74.7$\pm$4.3  & 73.0$\pm$3.7 & 75.9$\pm$6.7  \\
2-Bonsai&IRMv1 & 71.6$\pm$4.2 & 75.3$\pm$4.8    & 74.5$\pm$2.3 & 75.2$\pm$6.5  \\
2-Bonsai&vREx &  73.4$\pm$3.3 & 76.4$\pm$5.3   & 73.0$\pm$3.9 & 77.1$\pm$5.0  \\
\midrule
2-Bonsai-cf&ERM & 78.2$\pm$2.6 & 78.6$\pm$2.6 & 77.8$\pm$2.4 & 78.6$\pm$2.6 \\
2-Bonsai-cf&IRMv1 & 78.0$\pm$2.1 & 79.1$\pm$2.1  & 78.0$\pm$2.1 & 79.1$\pm$2.1 \\
2-Bonsai-cf&vREx & 77.9$\pm$2.7 & 79.5$\pm$2.7  & 78.0$\pm$2.6 & 79.7$\pm$2.4  \\
\bottomrule
    \end{tabular}
   
    \label{tab:rand_average}
\end{table*}

\end{document}